\documentclass{article} 
\usepackage[preprint]{colm2025_conference}

\usepackage{microtype}
\usepackage{hyperref}
\usepackage{url}
\usepackage{booktabs}
\usepackage{multicol}
\usepackage{multirow}
\usepackage{graphicx}
\usepackage{makecell}
\usepackage{algorithm}
\usepackage{algpseudocode}
\usepackage{amsmath}
\usepackage{wrapfig}
\usepackage{lineno}
\usepackage{pifont}
\usepackage{cleveref}

\crefname{figure}{Fig.}{Fig.}
\crefname{table}{Tab.}{Tab.}
\crefname{section}{Sec.}{Sec.}
\crefname{appendix}{App.}{App.}
\definecolor{darkblue}{rgb}{0, 0, 0.5}

\title{\texttt{MedReason}: Eliciting Factual Medical Reasoning Steps in LLMs via Knowledge Graphs
}

\author{
Juncheng Wu$^{1,\ast}$,Wenlong Deng$^{2,8,\ast}$,Xingxuan Li$^3$, Sheng Liu$^4$, Taomian Mi$^2$,\\
\textbf{Yifan Peng$^5$, Ziyang Xu$^6$, Yi Liu$^5$, Hyunjin Cho$^7$, Chang-In Choi$^9$, Yihan Cao$^{10}$,}\\
\textbf{Hui Ren$^{10}$, Xiang Li$^{10}$, Xiaoxiao Li$^{2,8,\dag}$, Yuyin Zhou$^{1,\dag}$} \\
$^1$UC Santa Cruz; $^2$University of British Columbia; $^3$Nanyang Technological University;\\
$^4$Stanford University;
$^5$Weill Cornell Medicine; $^6$NYU Langone Health; \\
$^7$Chungnam National University Sejong Hospital; $^8$Vector Institute;
\\
$^9$Pusan National University Hospital;
$^{10}$Massachusetts General Hospital;\\
\texttt{$^\ast$Equal Contribution, $^\dag$Corresponding author}\\
\url{https://github.com/UCSC-VLAA/MedReason}
}

\newcommand{\ours}{\texttt{MedReason}}

\begin{document}

\ifcolmsubmission
\linenumbers
\fi

\maketitle
\vspace{-10mm}
\begin{figure*}[h]
\centering
\includegraphics[width=0.88\linewidth]{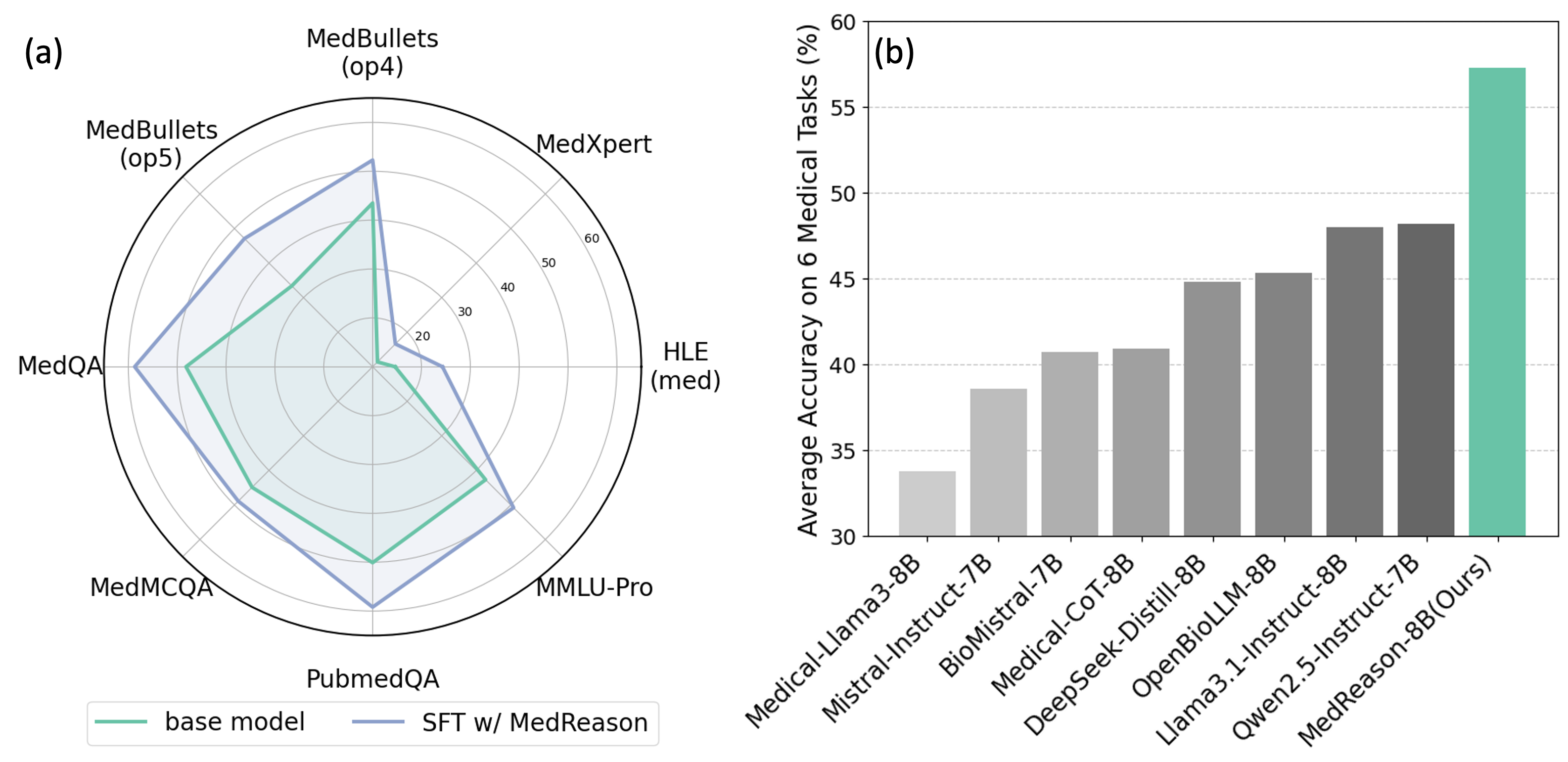} 
\vspace{-5mm}
\caption{\textbf{\texttt{MedReason}-8B significantly enhances medical reasoning capability in LLMs.} (a) SFT with \texttt{MedReason} consistently improves base LLMs across multiple datasets. (b) Our fine-tuned model achieves state-of-the-art performance among 7-8B LLMs.}
\label{fig:teaser}
\vspace{-2mm}
\end{figure*}

\begin{abstract}
Medical tasks such as diagnosis and treatment planning require precise and complex reasoning, particularly in life-critical domains. Unlike mathematical reasoning, medical reasoning demands meticulous, verifiable thought processes to ensure reliability and accuracy. 
 However, there is a notable lack of datasets that provide transparent, step-by-step reasoning to validate and enhance the medical reasoning ability of AI models. 
To bridge this gap, we introduce \textbf{\ours{}}, a large-scale high-quality medical reasoning dataset designed to enable faithful and explainable medical problem-solving in large language models (LLMs). We utilize a structured medical knowledge graph (KG) to convert clinical QA pairs into logical chains of reasoning, or ``thinking paths'', which trace connections from question elements to answers via relevant KG entities. Each path is validated for consistency with clinical logic and evidence-based medicine. Our pipeline generates detailed reasoning for various medical questions from 7 medical datasets, resulting in a dataset of \textbf{32,682} question-answer pairs, each with detailed, step-by-step explanations. 
Experiments demonstrate that fine-tuning with our dataset consistently boosts medical problem-solving capabilities, achieving significant gains of up to \textbf{7.7\%} for DeepSeek-Ditill-8B. Our top-performing model, \texttt{MedReason}-8B, outperforms the Huatuo-o1-8B, a state-of-the-art medical reasoning model, by up to \textbf{4.2\%} on the clinical benchmark MedBullets. We also engage medical professionals from diverse specialties to assess our dataset's quality, ensuring \texttt{MedReason} offers accurate and coherent medical reasoning.
\end{abstract}

\vspace{-0.1in}
\section{Introduction}
\vspace{-0.05in}
Recent advancements in Reasoning Large Language Models~\citep{xie2024preliminary,zhong2024evaluation,wang2024openr} highlight the remarkable effectiveness of utilizing \textit{Chain-of-Thought (CoT)} reasoning~\citep{huang2022towards,miao2024chain} prior to the final answers.
Although general-purpose reasoning models achieve parity with human performance in mathematical and coding tasks~\citep{guo2025deepseek,team2025kimi,jaech2024openai}, their applications in the medical domain have not been fully explored.
One of the key challenges is the scarcity of high-quality CoT data, which is essential for developing medical reasoning models.
Studies like s1~\citep{muennighoff2025s1} and LIMO~\citep{ye2025limo} have illustrated the crucial role of high-quality data in improving the LLMs' reasoning capability.
The limited scalability of this high-quality medical data has hindered the development of more powerful medical intelligence.

\begin{figure*}[t]
\centering
\includegraphics[width=\linewidth]{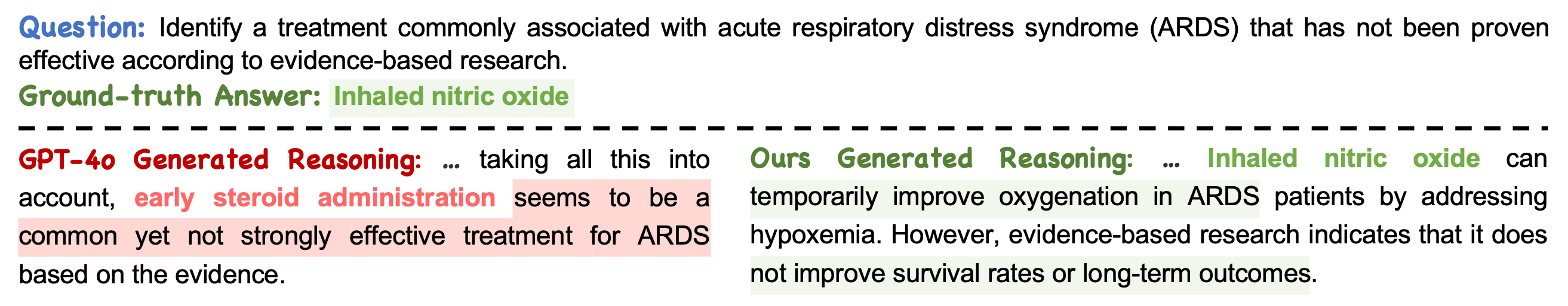} 
\vspace{-3mm}
\caption{An Example from Huatuo’s CoT Data, highlighting the factual error in the reasoning process generated by GPT-4o. In comparison, our generated reasoning leads to correct answer with accurate knowledge.}
\label{fig:error_case}
\vspace{-3mm}
\end{figure*}

One straightforward way to tackle this challenge is by distilling CoT data from open-source reasoning models~\citep{slam-distillation-from-r1,huang2024o1, min2024imitate,m1}. However, these methods often omit \textbf{quality filtering} (i.e. verifying whether the generated CoT data logically leads to the correct answer).
\cite{pang2025bolt} introduces BOLT, which integrates LLMs within a multi-agent framework to produce extensive CoT data while employing an outcome reward model to filter out low-quality reasoning traces.
Nevertheless, the proportion of medical-specific reasoning data remains limited, compromising the resulted model's clinical applicability.
Efforts like HuatuoGPT-o1~\citep{chen2024huatuogpt} aim to bridge this gap by generating medical CoT data using GPT-4o~\citep{hurst2024gpt}.
However, it is inevitable that general-purpose LLMs will generate responses that include factual errors.
As shown in~\cref{fig:error_case}, GPT-4o erroneously concludes that early administering steroids is not a highly effective treatment for ARDS, contradicting established findings~\citep{qadir2024update}.

\begin{figure*}[t]
\centering
\includegraphics[width=\linewidth]{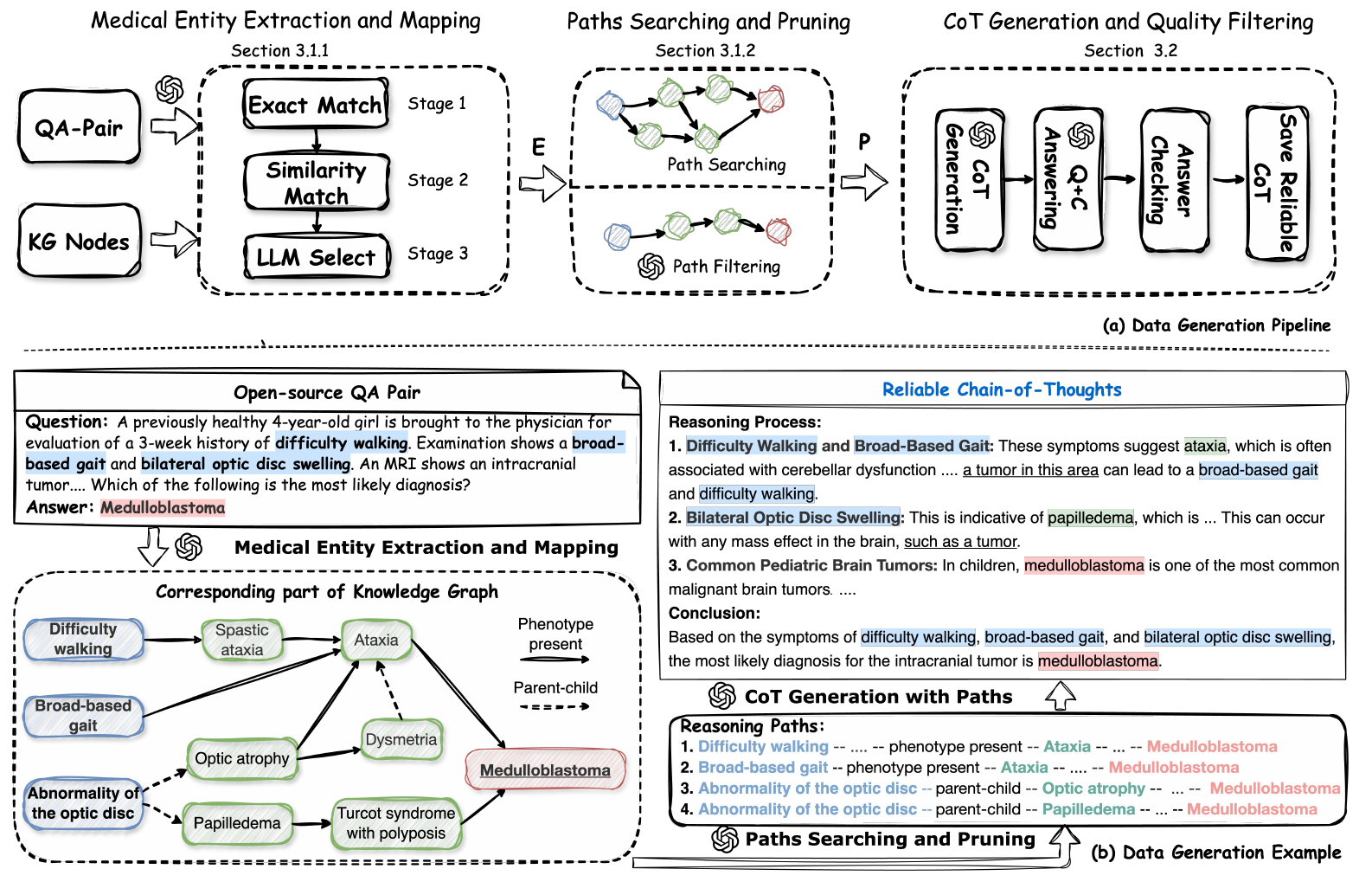} 
\vspace{-3mm}
\caption{\textbf{Overview of Our Data Generation Pipeline.} We first extract and map entities within each medical Q\&A pair, (see~\cref{sec:entity_extraction}). Next, we search and prune the reasoning paths between Q\&A entities in the KG (see~\cref{sec:path_searching}), which are utilized as factual guidance to construct the CoT data(see~\cref{sec:cot_generation}). Finally, we discard any generated CoT that can not lead to reach the correct answer to ensure the quality of our data (see~\cref{sec:quality_filtering}).}
\label{fig:example}
\vspace{-3mm}
\end{figure*}

In the medical domain, it is vital for the model to guarantee both the quality and rigorous factual guidance throughout every reasoning step, upholding medical reliability and clinical validity. 
In this paper, we argue that knowledge graph (KG) integration can provide \textbf{factual guidance} during CoT data generation, ensuring (1) logical coherence across all reasoning steps, and (2) clinical validity grounded in established medical knowledge. 
To achieve this, we propose a novel data generation pipeline that actively constrains the reasoning process to align with medical facts from KG, enhancing direct clinical utility. Specifically, our approach expands medical question-answering pairs into high-quality CoT data by searching reasoning paths from a high-quality medical KG~\citep{chandak2022building}, which serves as reliable medical knowledge sources.
We initially compile question-answering pairs from 7 medical datasets, encompassing general knowledge QA datasets as well as clinically challenging datasets which require complex reasoning.
As shown in~\cref{fig:example}, for each question-answering pair, we firstly prompt an LLM to extract entities from the question and answer components, then map these entities to corresponding nodes in our medical knowledge graph through either exact matching or LLM-based similarity selection (\cref{sec:entity_extraction}). 
We subsequently identify all reasoning paths connecting the question and answer entities within the knowledge graph, and instruct the LLM to prune reasoning paths that do not pertain to the current question (\cref{sec:path_searching}). 
Finally, the remaining reasoning paths serve as structural scaffolds to guide the LLM in generating medically grounded CoT explanations, enhancing interpretability and reasoning quality (\cref{sec:cot_generation}).

To ensure the quality of the generated CoT data,  we implement a verification step where an LLM answers each question using the generated reasoning path. We systematically eliminate any CoT samples that fail to produce correct answers, ensuring only logically sound and clinically valid reasoning paths are retained (\cref{sec:quality_filtering}). 
As shown in~\cref{tab:data_comparison}, our proposed generation pipeline yields 32,682 high-quality CoT samples with consistently improved reasoning quality across all evaluation metrics.

We assess the effectiveness of \ours{} through supervised fine-tuning (SFT) on 1) instruction fine-tuned models (LLaMA 3.1-Instruct-8B~\citep{grattafiori2024llama}, Mistral-Instruct-7B~\citep{jiang2023mistral7b}) and 2) medical reasoning specialists (Medical-CoT-8B~\citep{MedicalCOT}, DeepSeek-Distill-8B~\citep{guo2025deepseek}). 
Our extensive experimental evaluation on 7 QA benchmarks, encompassing 4 common medical benchmarks (MedQA~\citep{jin2021disease}, MedMCQA~\citep{pal2022medmcqa}, MMLU-Pro~\citep{wang2024mmlu}, and PubMedQA~\citep{jin2019pubmedqa}), and 3 challenging clinical benchmarks (MedBullets~\citep{chen2024benchmarking}, MedXpert~\citep{zuo2025medxpertqa}, and Humanity's Last Exam (HLE) ~\citep{phan2025humanity}) demonstrate the following key benefits of our generated CoT data: First, supervised fine-tuning (SFT) with our data yields consistent performance improvements across diverse base models and benchmarks. Notably, it enhances both instruction-tuned LLMs (\cref{tab:huatuo_comparison}) and specialized medical reasoning models (\cref{tab:reasoning_models}), with our best model achieving state-of-the-art performance among 7-8B parameter LLMs on challenging clinical benchmarks (\cref{tab:full_performance,fig:teaser}). Second, \ours{} produces higher-quality medical reasoning through knowledge-graph grounded generation, outperforming existing datasets~\citep{chen2024huatuogpt} in both automated metrics (\cref{tab:huatuo_comparison}) and expert evaluations conducted by physicians across seven clinical specialties (\cref{fig:selection}). Third, our approach enables superior clinical utility, as evidenced by side-by-side comparisons showing our model generates more factually precise and clinically supportive reasoning chains than competing approaches (\cref{fig:error_case} and \cref{fig:cases}).

\vspace{-3mm}
\begin{table}[t]
\setlength{\tabcolsep}{1.2mm}
  \centering
\scalebox{0.90}{
    \begin{tabular}{lcccc}
    \toprule
    \textbf{Data Source} &  \textbf{Quality Filtering} & \textbf{Medical Specific} & \textbf{Factual Guidance} & \textbf{Expert Checking} \\
    \midrule
    Distillation & \textcolor{red}{\textcolor{red}{\ding{55}}}    & \textcolor{red}{\ding{55}}    & \textcolor{red}{\ding{55}}    & \textcolor{red}{\ding{55}} \\
    BOLT  & \textcolor{green}{\ding{51}}   & \textcolor{red}{\ding{55}}    & \textcolor{red}{\ding{55}}    & \textcolor{red}{\ding{55}} \\
    Huatuo-o1 CoT & \textcolor{green}{\ding{51}}   & \textcolor{green}{\ding{51}}   & \textcolor{red}{\ding{55}}    & \textcolor{red}{\ding{55}} \\
    \texttt{MedReason}(ours) & \textcolor{green}{\ding{51}}   & \textcolor{green}{\ding{51}}   & \textcolor{green}{\ding{51}}   & \textcolor{green}{\ding{51}} \\
    \bottomrule
    \end{tabular}%
    }
    \caption{\textbf{Comparison between Chain-of-Thought data sources.} \ours{} provides high-quality medical CoT with factual guidance.  Medical experts from seven departments assess the generated CoT data sampled from \ours{}, further ensuring the quality of our data.}
  \label{tab:data_comparison}%
  \vspace{-6mm}
\end{table}%




\vspace{-0.1in}
\section{Related Works}
\vspace{-0.05in}
\paragraph{Reasoning with knowledge in LLMs.}
Recent reasoning large language models (LLMs) have demonstrated impressive performance in the math and coding domains~\citep{guo2025deepseek,team2025kimi,jaech2024openai,li2025system}, prompting the need for analogous development in the medical domain~\citep{goh2024large,lucas2024reasoning}.
However, training these models typically requires vast amounts of high-quality data that include intermediate reasoning steps~\citep{shao2024deepseekmathpushinglimitsmathematical, guo2025deepseek}.
Since manually annotating such data is not scalable, they are often distilled from more powerful LLMs.
This distillation process introduces unique challenges for tasks that demand factual knowledge, as LLMs can prone to generating hallucinations~\citep{huang2023surveyhallucinationlargelanguage}.
This issue is even more pronounced in the medical domain, where even state-of-the-art LLMs struggle to provide high-quality and accurate reasoning~\citep{griot2025large,chen2024cod,patel2005thinking}.
To address these challenges, this work introduces \ours{}, a medical reasoning dataset with high-quality CoT data designed to elicit factual-based and interpretable medical reasoning within LLMs.

\paragraph{LLM-Distilled Medical Datasets.}

Pre-training and fine-tuning medical LLMs demand extensive and high-quality datasets. 
Earlier research has primarily focused on gathering instruction-tuning data to imbue general domain LLMs with medical expertise~\citep{xie2024me,li2023llava}.
Recently, to improve the medical reasoning capability of LLMs, \citet{chen2024huatuogpt} introduced a medical CoT dataset by leveraging GPT-4o~\citep{hurst2024gpt} for strategy-based retrieval, which yielded 20K question-answer pairs with complex CoT data.
However, directly employing general domain LLMs to generate CoT data poses challenges in maintaining the integrity of medical knowledge at every reasoning step.
Inspired by \cite{xie2024medtrinity}, which obtains dependable information to assist in producing multigranular captions for medical images, our work employs a medical knowledge graph~\cite{chandak2022building} to provide factual guidance during the generation of medical CoT data.

\vspace{-0.1in}
\section{Method}
\vspace{-0.05in}
\subsection{Retrieving Reasoning Paths from Knowledge Graph}
In this section, we detail the process of retrieving reasoning paths from the knowledge graph (KG).
In the data generation process, we define the Language Model as $LLM$ and the knowledge graph as $G$. We utilize OpenAI GPT-4o~\citep{hurst2024gpt} as $LLM$ and employ PrimeKG~\citep{chandak2022building} as our knowledge base. 
\vspace{-0.05in}
\subsubsection{Medical Entity Extraction and Mapping}\label{sec:entity_extraction}
\vspace{-0.05in}
As shown in \cref{fig:example} (b), given a question \( Q \) and its corresponding answer \( A \), we first utilize the Language Model $LLM$ to identify the medical entities present in \( Q \) and \( A \). This results in the extracted entity sets \( \{e_i^Q\}_{i \in [n]} \) and \( \{e_j^A\}_{j \in [m]} \), where $n$ and $m$ denote the number of entities in $Q$ and $A$, respectively. These entities are then mapped to the corresponding nodes in the knowledge graph \( G \) through a three-step mapping process,  as shown in Fig~\ref{fig:example} (a). First, a text embedding model is used to encode each entity \( e \in E \) and compute its similarity with the node embeddings in the \( G \). This generates a ranked list of candidate matches, from which we then extract the Top-\( K \) most similar entities to form a candidate set \( S \). Thirdly, we select the final entity from the Top-\( K \) entities, following the three matching stages: 
\\
\noindent \textbf{Stage 1 (Exact Match):} The algorithm iterates over \( S \) and checks for an exact match with \( e \). If an exact match is found $e \in S$, the corresponding node is selected.  
\\
\textbf{Stage 2 (Similarity Match):} If an exact match is not found and the top similarity score exceeds a predefined threshold \( \tau \) (set to 0.85 in our case), the most similar entity from \( S \) is selected.  
\begin{align}
    \hat{e} = \arg\max_{s_k \in S} \cos(e,s_k), ~\text{if}  ~\cos(e,s_k) > \tau
\end{align}
\\
\textbf{Stage 3 (LLM-based Selection):} If no suitable candidate is found in the above two stages, we instruct the LLM to analyze the question-answer context and the entity name to determine the most relevant node from \( S \). The selection prompt $I_{\text{select}}$ is demonstrated in \cref{fig:rela_prompt} in the Appendix.
\begin{align}
     \hat{e} &= LLM\left(S, Q, A\mid I_{\text{select}} \right),
\end{align}
Finally, we derive mapped entity sets from the graph, denoted as \( \{\hat{e}_i^Q\}_{i \in [n]} \) and \( \{\hat{e}_j^A\}_{j \in [m]} \), respectively.  We detail the algorithm in Appendix Algorithm~\ref{alg:entity_mapping}. As illustrated in Fig.~\ref{fig:example}(b), besides the entities \textit{difficulty walking} and \textit{broad-based gait} which exactly match graph nodes, the entity \textit{bilateral optic disc swelling} is mapped to a similar concept, \textit{Abnormality of the optic disc}, in the knowledge graph.
\vspace{-0.05in}
\subsubsection{Paths Searching and Pruning}
\label{sec:path_searching}
\vspace{-0.05in}
Given the mapped entities \( \{\hat{e}_i^Q\}_{i \in [n]} \) from the question and \( \{\hat{e}_j^A\}_{j \in [m]} \) from the answer, our goal is to identify reasoning paths that logically connect the question to its corresponding answer.
These paths will later serve as guidance for CoT generation (\cref{sec:cot_generation}), ensuring that every reasoning step (1) originates from authoritative medical knowledge, and (2) maintains factual consistency with the KG.
Specifically, we determine the shortest paths to avoid overthinking~\citep{luo2025o1,chen2024not} and maintain concise reasoning for each pair of question-answer entities \( \{ \hat{e}_i^Q,\hat{e}_j^A \} \), which identify the most immediate correlations. This set of shortest paths for the node pair is represented as $\tilde{P}_{i,j}$. We demonstrate examples of paths in \cref{fig:example} (b), where the entities in question (\textcolor{blue}{blue} nodes) are connected to the answer entity (\textcolor{red}{red} node). However, there may be a significant number of paths of the same length linking $\hat{e}_i^Q$ to $\hat{e}_j^A$ within the KG $G$.
To ensure that we can retrieve the reasoning paths that are correlated to our question, we employ the $LLM$ to prune the irrelevant paths as shown in \cref{fig:example} (a). 
In particular, we provide the shortest paths set $\tilde{P}_{i,j}$ and the question $Q$ to the $LLM$, which is prompted to select $K$ paths that are most correlated to the question.
In summary, for the node pair \( \{ \hat{e}_i^Q,\hat{e}_j^A \} \), the reasoning paths searching and filtering process is denoted as:
\begin{equation}
\label{eq:path_searching}
\begin{aligned}
    \{\tilde{P}_{i,j}^k\}_{k\in[K]}&=\texttt{shortest\_paths}\left(\hat{e}_i^Q, \hat{e}_j^A, G \right), \\
P_{i,j}^k &= LLM\left(\tilde{P}_{i,j}^k, Q \mid I_{\text{prune}} \right), \quad k\in [K]
\end{aligned}
\end{equation}
where $I_{\text{prune}}$ is the path pruning prompt, and we set $K=3$ during data generation. More details can be found in Appendix~\cref{fig:purne_prompt}. Finally, we aggregate all identified reasoning paths across the question-answer pairs, forming the complete set \( \mathcal{P} = \{P_{i,j}^k\}_{i \in [n], j \in [m], k \in [K]} \), for CoT generation.
As illustrated in~\cref{fig:example} (b), we explore various reasoning paths that connect question and answer entities within the corresponding subgraph of the KG. 
After filtering out the irrelevant paths, the pruned paths effectively identify the link between symptoms like 'Difficulty walking' and the diagnosis 'Medulloblastoma', uncovering the critical intermediary disease 'Ataxia'. 
\vspace{-0.05in}
\subsection{CoT Generation and Quality Filtering}
\vspace{-0.05in}
\paragraph{CoT Generation with Reasoning Paths}
\label{sec:cot_generation}
Utilizing step-by-step reasoning paths in $\mathcal{P}$ as guidance, we are able to distill the reliable knowledge from the off-the-shelf KG into our CoT data.
To achieve this, we prompt the $LLM$ to analyze the given reasoning paths and elaborate on the relevant ones to formulate medically grounded CoT explanations of the response, represented as:
\begin{equation}
\label{eq:cot_generation}
    C = LLM\left(Q,A,\mathcal{P} \mid I_{\text{gen}}\right),
\end{equation}
where $I_{\text{gen}}$ represents our carefully designed generation prompt (see Appendix~\cref{fig:gen_prompt}). As illustrated in~\cref{fig:example}b, our approach produces clinically grounded reasoning chains - for instance, beginning with symptom analysis, progressing through pathological deduction (e.g., tumor identification), and ultimately concluding the final diagnosis. Each step maintains direct alignment with the KG-derived evidence in $\mathcal{P}$, ensuring both factual accuracy and clinical relevance.
\\
\textbf{Quality Filtering}\label{sec:qf}
\label{sec:quality_filtering}
Lastly, to ensure the quality of the generated data, we design a simple quality filtering strategy to filter out low-quality CoT data. 
Specifically, for each generated CoT $C$, we prompt the LLM to produce an answer $\hat{A}$ using \emph{only} the information contained in $C$ (see details in Appendix~\cref{fig:eval_prompt}). 
The answer generation process can be denoted as:
\begin{equation}
\label{eq:answer_generation}
    \hat{A}=LLM\left(Q, C\mid I_{\text{eval}} \right),
\end{equation}
where $I_{\text{eval}}$ denotes the prompt for generation. Subsequently, as illustrated in~\cref{fig:example} (a), we compare $\hat{A}$ with the original ground-truth answer $A$.
We apply this quality filtering strategy to all 45K generated samples, retaining only those CoT instances (32K) that yield correct answers, to ensure both logical validity and factual accuracy in our final dataset.
\vspace{-0.1in}
\section{Experiment}


\vspace{-0.05in}
\subsection{Experimental Setup} 
\vspace{-0.02in}
\paragraph{Data Collection and Preprocessing.}
Our data pipeline generates CoT reasoning for Question-Answering (QA) pairs. To curate medical QA pairs, we gather datasets such as MedQA~\citep{jin2021disease}, MedMCQA~\citep{pal2022medmcqa}, PubmedQA~\citep{jin2019pubmedqa}, MMLU~\citep{hendrycks2020measuring}, HuatuoGPT-o1~\citep{chen2024huatuogpt}, MedXpert~\citep{zuo2025medxpertqa}, and Humanity's Last Exam (HLE)~\citep{phan2025humanity}. To prevent data leakage, we exclusively use the training set from each dataset for CoT data generation, culminating in a total of 55K QA pairs. We exclude the QA pairs that cannot produce CoT data using our pipeline due to the absence of entities in either the question or answer, such as when the answer is merely a number, ultimately leading to 45K QA pairs for CoT data generation. Additional statistics for the generated and quality-filtered datasets are provided in Table~\ref{tab:data_statistics} in the Appendix.
\\
\textbf{Baseline Models.} To evaluate the effectiveness of our dataset across various base models, we select multiple 7-8B models as baselines and conduct supervised fine-tuning (SFT) on \ours{}. Specifically, we fine-tune two representative instruction-tuned models: LLaMA 3.1-Instruct-8B~\citep{grattafiori2024llama} and Mistral-Instruct-7B~\citep{jiang2023mistral7b}. Following~\citep{chen2024huatuogpt}, we train these models for three epochs using a learning rate of 5e-6 and a batch size of 128, employing DeepSpeed-ZeRO stage 3~\citep{rajbhandari2020zero}. To further assess the impact of our dataset on reasoning models, we also fine-tune Medical-CoT-8B~\citep{MedicalCOT}, DeepSeek-Distill-8B~\cite{guo2025deepseek}, and Huatuo-o1-RL-8B~\citep{chen2024huatuogpt} using the same hyperparameter settings.
\\
Finally, in~\cref{tab:full_performance}, we benchmark our model against three categories of models: (1) General LLMs, including LLaMA 3.1-Instruct-8B~\citep{grattafiori2024llama}, Mistral-Instruct-7B~\citep{jiang2023mistral7b}, and Qwen-Instruct-7B~\citep{yang2024qwen2}; (2) Medical-Specific LLMs, such as Medical-Llama~\citep{qiu2024towards}, OpenBioLLM~\citep{OpenBioLLMs}, Huatuo-o1-SFT~\citep{huang2024o1}, and BioMistral~\cite{labrak2024biomistral}; and (3) Medical Reasoning Models, including Medical-CoT~\citep{MedicalCOT} and Huatuo-o1-RL~\citep{huang2024o1}.
\\
\textbf{Benchmarks.} We evaluate on standard medical benchmarks: MedQA (USMLE test set)~\citep{jin2021disease} , MedMCQA (validation set)~\citep{pal2022medmcqa}, health and biology tracks of MMLU-Pro~\citep{wang2024mmlu}, and PubMedQA (test set)~\citep{jin2019pubmedqa}. Additionally, we evaluated the medical sections of some challenging LLM benchmarks, including the clinical expertise benchmark MedBullets~\citep{chen2024benchmarking}, expert-level medical knowledge and advanced reasoning MedXpert~\citep{zuo2025medxpertqa}, and the most recent medical questions in Humanity's Last Exam (HLE) ~\citep{phan2025humanity}.

\begin{table}[t]
\setlength{\tabcolsep}{0.5mm}
  \centering
    \begin{tabular}{lcccccc}
    \toprule
    \multicolumn{1}{l}{\multirow{2}[4]{*}{\textbf{Benchmarks}}} & \multicolumn{3}{c}{\textbf{Llama3.1-Instruct-8B}} & \multicolumn{3}{c}{\textbf{Mistral-Instruct-7B}} \\
\cmidrule{2-7}          & \textbf{original}  & \textbf{w/ huatuo CoT} & \textbf{w/ ours} & \textbf{original}  & \textbf{w/ huatuo CoT} & \textbf{w/ ours} \\
    \midrule
    MedQA & 58.7  & \textbf{70.2\textcolor{red}{~(+11.5)}} & 68.4\textcolor{red}{~(+9.7)} & 48.2  & \textbf{59.9\textcolor{red}{~(+11.7)}} & 58.7\textcolor{red}{~(+10.5)} \\
    MedMCQA & 56.0  & \textbf{58.2\textcolor{red}{~(+2.2)}} & 57.5\textcolor{red}{~(+1.5)} & 44.9  & 46.9\textcolor{red}{~(+2.0)} & \textbf{48.9\textcolor{red}{~(+4.0)}} \\
    PubmedQA & 75.2  & 76.1\textcolor{red}{~(+0.9)} & \textbf{77.6\textcolor{red}{~(+2.4)}} & 50.1  & 57.5\textcolor{red}{~(+7.4)} & \textbf{59.2\textcolor{red}{~(+9.1)}} \\
    MMLU-Pro & 58.2  & 59.9\textcolor{red}{~(+1.7)} & \textbf{63.1\textcolor{red}{~(+4.9)}} & 42.7  & 47.6\textcolor{red}{~(+4.9)} & \textbf{50.8\textcolor{red}{~(+8.1)}} \\
    MedBullets(op4)  & 48.7  & 53.3\textcolor{red}{~(+4.6)} & \textbf{57.5\textcolor{red}{~(+8.8)}} & 43.5  & 50.0\textcolor{red}{~(+6.5)} & \textbf{52.3\textcolor{red}{~(+8.8)}} \\
    MedBullets(op5)  & 42.5  & 49.7\textcolor{red}{~(+7.2)} & \textbf{52.3\textcolor{red}{~(+9.8)}} & 33.4  & 46.1\textcolor{red}{~(+12.7)} & \textbf{47.1\textcolor{red}{~(+13.7)}} \\
    MedXpert & 13.2  & \textbf{17.3\textcolor{red}{~(+4.1)}} & 16.4\textcolor{red}{~(+3.2)} & 11.4  & 14.4\textcolor{red}{~(+3.0)} & \textbf{16.6\textcolor{red}{~(+5.2)}} \\
    HLE (med) & 13.6  & 14.6\textcolor{red}{~(+1.0)} & \textbf{16.5\textcolor{red}{~(+2.9)}} & 14.6  & 14.6\textcolor{gray}{~(+0.0)} & \textbf{24.3\textcolor{red}{~(+9.7)}} \\
    \midrule
    Avg   & 45.8  & 49.9\textcolor{red}{~(+4.1)} & \textbf{51.2\textcolor{red}{~(+5.4)}} & 36.1  & 42.1\textcolor{red}{~(+6.0)} & \textbf{44.7\textcolor{red}{~(+8.6)}} \\
    \bottomrule
    \end{tabular}%
    \caption{\textbf{Results of instruction-tuned LLMs fine-tuned with Huatuo complex CoT and \ours{} (Ours).} Integrating our high-quality reasoning data into supervised fine-tuning significantly improves model performance across various benchmarks and model types.}
  \label{tab:huatuo_comparison}%
  \vspace{-2mm}
\end{table}%

\begin{table}[t]
\setlength{\tabcolsep}{0.4mm}
  \centering
\scalebox{0.95}{
    \begin{tabular}{llccccc}
    \toprule
    \multirow{2}[4]{*}{\textbf{Base Model}} & \multirow{2}[4]{*}{\textbf{Data}} & \multicolumn{5}{c}{\textbf{\small{Clinical Challenging Datasets}}} \\
\cmidrule{3-7}          &       & \textbf{\small{MedBullets(op4)}}  & \textbf{\small{MedBullets(op5)}}  & \textbf{\small{MedXpert}} & \textbf{\small{HLE (med)}} & \textbf{\small{Avg}} \\
    \midrule
    \multirow{2}[2]{*}{\small{Medical-CoT-8B}} & \small{original} & 39.3  & 34.1  & 12.6  & 15.5  & 25.4 \\
          & \small{w/ ours} & \textbf{49.0\textcolor{red}{~(+9.7)}} & \textbf{41.9\textcolor{red}{~(+7.8)}} & \textbf{14.2\textcolor{red}{~(+1.6)}} & \textbf{17.5\textcolor{red}{~(+2.0)}} & \textbf{30.6\textcolor{red}{~(+5.3)}} \\
    \midrule
    \multirow{2}[2]{*}{\small{DeepSeek-Distill-8B}} & \small{original} & 41.9  & 35.1  & 13.5  & 11.7  & 25.5 \\
          & \small{w/ ours} & \textbf{53.6\textcolor{red}{~(+11.7)}} & \textbf{49.0\textcolor{red}{~(+14.0)}} & \textbf{15.9\textcolor{red}{~(+2.4)}} & \textbf{14.6\textcolor{red}{~(+2.9)}} & \textbf{33.3\textcolor{red}{~(+7.7)}} \\
    \midrule
    \midrule
    \multirow{2}[2]{*}{\textbf{Base Mode}l} & \multirow{2}[2]{*}{\textbf{Data}} & \multicolumn{5}{c}{\textbf{\small{Common MedicalQA Datasets}}} \\
    \cmidrule{3-7} &       & \textbf{\small{MedQA}} & \textbf{\small{MedMCQA}} & \textbf{\small{PubmedQA}} & \textbf{\small{MMLU-Pro}} & \textbf{\small{Avg}} \\
    \midrule
    \multirow{2}[2]{*}{\small{Medical-CoT-8B}} & \small{original} & 49.0  & 42.6  & 68.0  & 48.7  & 52.1 \\
          & \small{w/ ours} & \textbf{58.0\textcolor{red}{~(+9.0)}} & \textbf{46.6\textcolor{red}{~(+4.0)}} & \textbf{74.6\textcolor{red}{~(+6.6)}} & \textbf{50.4\textcolor{red}{~(+1.7)}} & \textbf{57.4\textcolor{red}{~(+5.3)}} \\
    \midrule
    \multirow{2}[2]{*}{\small{DeepSeek-Distill-8B}} & \small{original} & 55.4  & 49.0  & \textbf{73.9} & 53.8  & 58.0 \\
          & \small{w/ ours} & \textbf{63.7\textcolor{red}{~(+8.3)}} & \textbf{51.8\textcolor{red}{~(+2.8)}} & 73.0\textcolor{blue}{~(-0.9)} & \textbf{57.5\textcolor{red}{~(+3.8)}} & \textbf{61.5\textcolor{red}{~(+3.5)}} \\
    \bottomrule
    \end{tabular}%
}
  \caption{\textbf{Fine-tuning with our data further enhances reasoning LLMs.} We perform supervised fine-tuning on reasoning models in both general and medical domains, with our data consistently improving performance by providing high-quality medical knowledge.}
  \label{tab:reasoning_models}%
  \vspace{-5mm}
\end{table}%

\vspace{-0.1in}
\subsection{Results}
\vspace{-0.05in}
\paragraph{\ours{} on Instruction Fine-tuned Model.} In this section, we showcase the enhancement of Instruction Finetuned Models using \ours{}. \cref{tab:huatuo_comparison} displays the accuracy (\%) of Llama3.1-Instruct-8B and Mistral-Instruct-7B across various medical benchmarks. The model finetuned on \ours{} (\textbf{w/ ours}) consistently outperforms both the base models and the Huatuo CoT data~\citep{chen2024huatuogpt} finetuned model.
\\
For Llama3.1-Instruct-8B, finetuning using \ours{} improves the average accuracy from \textbf{45.8\%} to \textbf{51.2\%} (\textbf{+5.4\%}), surpassing the \textbf{+4.1\%} gain achieved with Huatuo CoT data. In the case of Mistral-Instruct-7B, the improvement is even more substantial, rising from \textbf{36.1\%} to \textbf{44.7\%} (\textbf{+8.6\%}), exceeding Huatuo CoT data's \textbf{+6.0\%} gain. Our method demonstrates consistent and substantial improvements, particularly in complex reasoning tasks (HLE) and specialized clinical benchmarks (MedBullets, MedXpert). 
\\
\textbf{\ours{} on Reasoning Models.}
We then showcase the improvements in reasoning models achieved through fine-tuning with our \ours{} dataset. As shown in \cref{tab:reasoning_models}, fine-tuning with our dataset (w/ ours) significantly enhances performance across both clinical and general medical question-answering tasks compared to the original model without fine-tuning. On challenging clinical datasets, Medical-CoT-8B achieves an average gain of $5.3\%$, while DeepSeek-Distill-8B demonstrates an even greater improvement of $7.7\%$. Similarly, for general medical QA datasets, Medical-CoT-8B improves by $5.3\%$ on average, and DeepSeek-Distill-8B gains $3.5\%$. These results highlight the effectiveness of our KG-driven dataset \ours{} in enhancing reasoning LLMs by integrating factual guided medical knowledge.
\\
\textbf{\ours{} achieves state-of-the-art on 7B-8B LLMs.} Finally, we obtain the state-of-the-art model by fine-tuning Huatuo-o1-RL-8B with \ours{}, denoting it as \ours{}-8B. The results in \cref{tab:full_performance} show that \ours{}-8B, fine-tuned using \ours{}, outperforms all other models across five evaluation datasets, achieving the highest average score of $\textbf{57.3\%}$. It surpasses its base model, Huatuo-o1-RL-8B, by $1.4\%$, demonstrating the effectiveness of \ours{} fine-tuning. Notably, it demonstrates even greater improvements on challenging reasoning tasks, outperforming Huatuo-o1-RL-8B by $4.2\%$ on MedBullets (op5) and $2.3\%$ on MedXpert. Compared to other strong baselines, such as OpenBioLLM-8B and DeepSeek-Distill-8B, \texttt{MedReason}-8B demonstrates substantial performance gains of approximately $12\%$. These results confirm that our KG-guided CoT data enhances reasoning capabilities, establishing a new state-of-the-art for medical QA tasks.
\begin{table}[t]
\setlength{\tabcolsep}{0.5mm}
  \centering
\scalebox{0.90}{
    \begin{tabular}{lccccccc}
    \toprule
    \textbf{Model} & \makecell{\textbf{MedBullets} \\ \textbf{(op4)}} & \makecell{\textbf{MedBullets} \\ \textbf{(op5)}} & \textbf{MedXpert} & \textbf{MedQA} & \textbf{MedMCQA} & \textbf{PubmedQA} & \textbf{Avg} \\
    \midrule
    Llama3.1-Instruct-8B & 43.2  & 40.9  & 14.3  & 58.7  & 56.0  & 75.2  & 48.0 \\
    Qwen2.5-Instruct-7B & 50.0  & 41.6  & 12.6  & 57.0  & 55.6  & 72.7  & 48.2 \\
    Mistral-Instruct-7B & 43.5  & 33.4  & 11.4  & 48.2  & 44.9  & 50.1  & 38.6 \\
    \midrule
    Medical-Llama3-8B & 33.4  & 25.3  & 9.0   & 40.3  & 46.8  & 48.0  & 33.8 \\
    OpenBioLLM-8B & 39.2  & 35.7  & 10.7  & 57.7  & 54.1  & 74.1  & 45.3 \\
    BioMistral-7B & 46.4  & 33.1  & 12.4  & 45.0  & 40.2  & 66.9  & 40.7 \\
    \midrule
    Medical-CoT-8B & 39.3  & 34.1  & 12.6  & 49.0  & 42.6  & 68.0  & 40.9 \\
    DeepSeek-Distill-8B & 41.9  & 35.1  & 13.5  & 55.4  & 49.0  & 73.9  & 44.8 \\
    Huatuo-o1-SFT-8B & 53.3  & 49.7  & 17.3  & 70.2  & 58.2  & 76.1  & 54.1 \\
    Huatuo-o1-RL-8B & 55.2  & 51.3  & 16.7  & \textbf{72.6} & 60.4  & 79.2  & 55.9 \\
    \midrule
    \ours{}-8B (\textbf{ours})  & \textbf{57.5} & \textbf{55.5} & \textbf{19.0} & 71.8  & \textbf{60.7} & \textbf{79.4} & \textbf{57.3} \\
    \bottomrule
    \end{tabular}
    }
    \caption{\textbf{Comparison across various medical benchmarks.} Our best model achieves state-of-the-art performance comparing with 7B-8B LLMs.}
    \label{tab:full_performance}
\end{table}

\begin{table}[t]
\setlength{\tabcolsep}{0.8mm}
    \centering
\scalebox{0.95}{
    \begin{tabular}{lccccccc}
    \toprule
    \makecell[l]{\textbf{Quality} \\ \textbf{Filtering}} & \textbf{MedQA} & \textbf{MedMCQA} & \textbf{PubmedQA} & \makecell{\textbf{MedBullets} \\ \textbf{(op4)}} & \makecell{\textbf{MedBullets} \\ \textbf{(op5)}} & \textbf{MedXpert} & \textbf{Avg} \\
    \midrule
    w/o    & 0.669 & 0.569 & 0.737 & 0.571 & 0.510 & \textbf{0.178} & 0.539 \\
    w/   & \textbf{0.684} & \textbf{0.575} & \textbf{0.776} & \textbf{0.575} & \textbf{0.523} & 0.164 & \textbf{0.550} \\
    \bottomrule
    \end{tabular}%
    }
    \caption{\textbf{The ablation study on quality filtering.} The results demonstrate that maintaining high-quality (correct) generated CoT consistently enhanced overall performance.}
    \label{tab:qf_perform}
    \vspace{-5mm}
\end{table}

\subsection{Ablation Study and Expert Verification}
\begin{figure*}[t]
\centering
\includegraphics[width=0.98\linewidth]{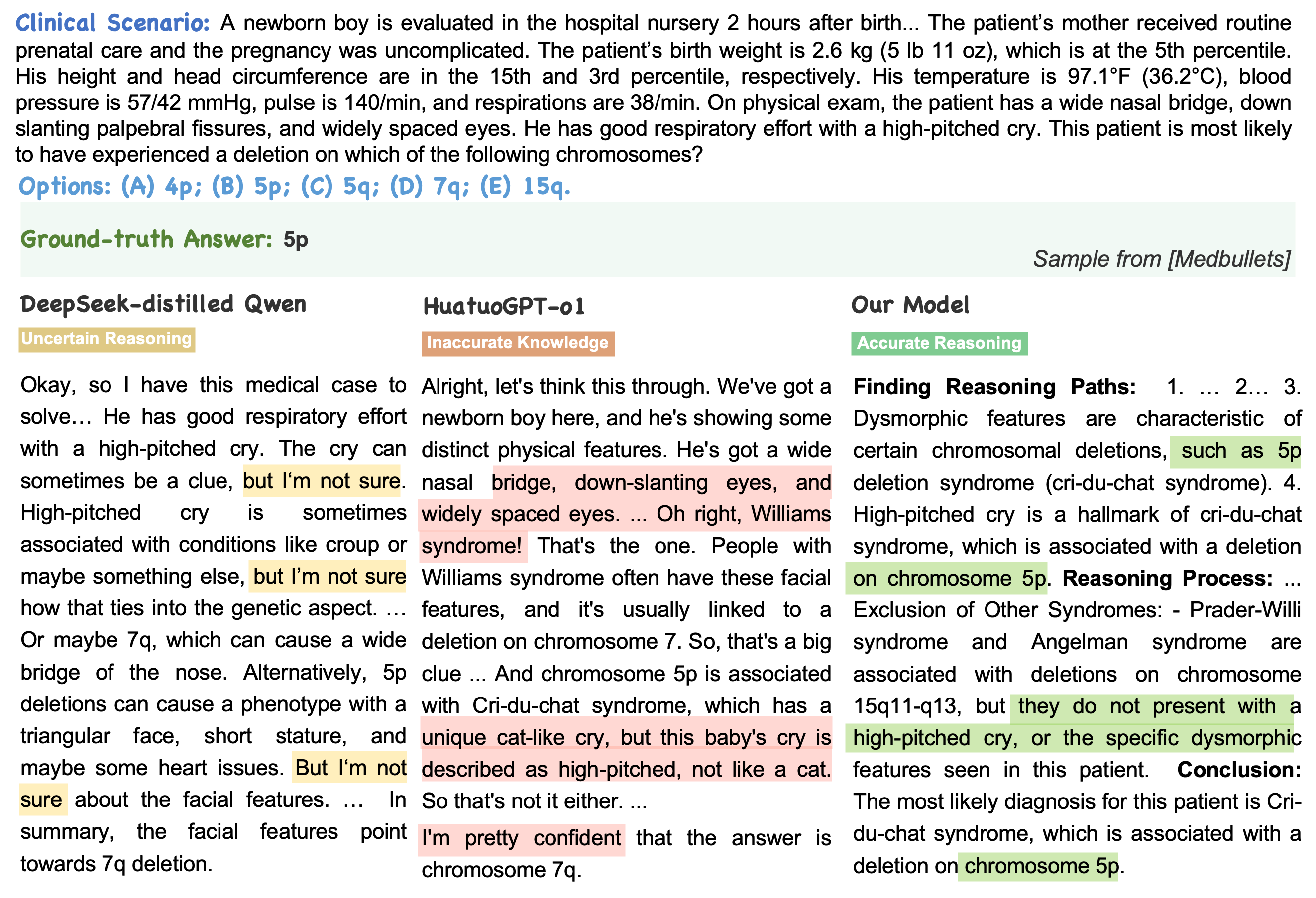} 
\vspace{-4mm}
\caption{Case Study on Medbullets Benchmark. Some part of reasoning is omitted due to the space limitation. Our model generates accurate reasoning with reliable knowledge.}
\label{fig:cases}
\vspace{-3mm}
\end{figure*}

\paragraph{Effect of Quality Filtering.} As outlined in Section~\ref{sec:qf}, we implement quality filtering to ensure the generated CoT data effectively guides a LLM to produce correct answers. To evaluate its impact, we perform an ablation study using the LLama3.1-Instruct-8B model. Table~\ref{tab:qf_perform} illustrates that quality filtering enhances performance across the majority of medical datasets, enhancing the average score by $1.1\%$. This further underscores the crucial role of CoT data quality in enhancing the reasoning capabilities of LLMs in medical applications.
\\
\textbf{CoT Case study.} To showcase the effectiveness of our dataset, we analyze a challenging medical case from MedBullets, comparing its performance against two other reasoning models—DeepSeek-distilled and HuatuoGPT-o1—in \cref{fig:cases}. Our dataset accurately diagnosed Cri-du-Chat syndrome, linking the patient’s symptoms to a chromosome 5p deletion, consistent with the ground-truth answer, thus validating its reasoning precision.  In contrast, DeepSeek-distilled model frequently responded with \textit{I'm not sure}, as indicated by the yellow shadow in Fig.~\ref{fig:cases}, reflecting uncertainty and a lack of clarity, which makes it unsuitable for assisting doctors. While HuatuoGPT-o1 provided a more confident response, it relied on incorrect knowledge, as indicated by the red shadow, leading to an inaccurate diagnosis.
 \begin{figure*}[t]
\centering
\includegraphics[width=\linewidth]{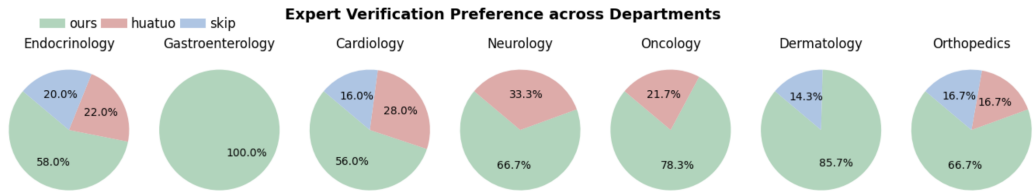} 
\caption{\textbf{Expert verification on CoT data quality} across seven medical specialties. Doctors were shown CoT data from both our dataset (\textcolor{green}{green}) and Huatuo-Complex CoT (\textcolor{red}{red}) and asked to pick the higher-quality explanation or skip (\textcolor{blue}{blue}) if neither was clearly superior. Each pie indicates the proportion of doctors in that specialty who selected our data, Huatuo-Complex CoT, or skipped. Our CoT data was consistently favored across all the specialties. }
\label{fig:selection}
\vspace{-5mm}
\end{figure*}
\\
\textbf{Expert Verification.}  To further evaluate the medical accuracy and clarity of CoT data within \ours{}, we engaged domain experts from seven different medical specialties. Each expert was given two anonymized CoT data—one produced by \ours{} and the other by HuatuoGPT-o1—and asked to choose which explanation, if any, was more accurate and easier to understand. Experts could also skip if they found both CoT data to be equally good or equally inadequate. For each specialty, we evaluated responses to over 25 randomly sampled questions.
\\
\cref{fig:selection} presents the results, illustrating the proportion of high-quality CoT data sources verified by licensed clinicians. As shown, specialists across all departments preferred \ours{}, with Gastroenterology experts unanimously selecting \ours{} ($100\%$), and Dermatology and Oncology also demonstrated a strong preference (over $80\%$). These findings suggest that \ours{} consistently provides more medically precise and coherent reasoning across multiple specialties, underscoring its potential for real-world clinical applications.
\section{Conclusion}
This paper presents a scalable, knowledge-guided pipeline for automatically generating high-quality medical Chain-of-Thought (CoT) data. By leveraging structured knowledge graphs to anchor the reasoning process, our method produces medically grounded and interpretable explanations that enhance the clinical validity of LLM-generated reasoning.
Experiments on instruction-tuned and reasoning-specialized LLMs demonstrate consistent improvements across medical benchmarks, particularly in complex clinical scenarios, while expert evaluations confirm superior reasoning quality over prior methods.  
We hope our work can encourage future exploration on clinically valid reasoning to advance trustworthy medical AI.


\bibliography{colm2025_conference}
\bibliographystyle{colm2025_conference}

\newpage
\appendix
\section{Appendix}

\subsection{OpenAI API Usage}
In this work, we employ GPT-4o from Azure in our data generation pipeline. The version for GPT-4o is 'gpt-4o-0806-nofilter-global', and the API version is '2024-12-01-preview'. Total API usage is about \$3,600.

\subsection{Data Statistics}
The following table demonstrates the specification of the data distribution of \ours{} across various source datasets.
In the case of Humanity's Last Exam (HLE)~\citep{phan2025humanity} and MedXpert~\citep{zuo2025medxpertqa}, we allocated 57 and 666 data samples, respectively, for training and reserved the rest for testing. For other datasets, we strictly use only the training data for data generation to avoid any potential data leakage.
\begin{table}[htbp]
\setlength{\tabcolsep}{0.8mm}
  \centering
\scalebox{0.90}{
    \begin{tabular}{lcccccccc}
    \toprule
    \textbf{Datasets} & \textbf{MedQA} & \textbf{MedMCQA} & \textbf{PubmedQA} & \textbf{MMLU} & \textbf{MedXpert} & \textbf{Huatuo} & \textbf{HLE} & \textbf{Total} \\
    \midrule
    Raw   & 9595  & 9131  & 24826 & 1089  & 1000  & 9271  & 159   & 55071 \\
    Generated & 8528  & 7598  & 20613 & 893   & 951   & 7010  & 132   & 45725 \\
    Quality Filtered & 8016  & 6197  & 10444 & 827   & 666   & 6475  & 57    & 32682 \\
    \bottomrule
    \end{tabular}%
}
  \caption{\textbf{Statistics of the collected QA dataset (Raw), Generated dataset, and final Quality-Filtered dataset.}}
  \label{tab:data_statistics}%
\end{table}%

\subsection{Prompts}
\begin{figure*}[h]
\centering
\includegraphics[width=\linewidth]{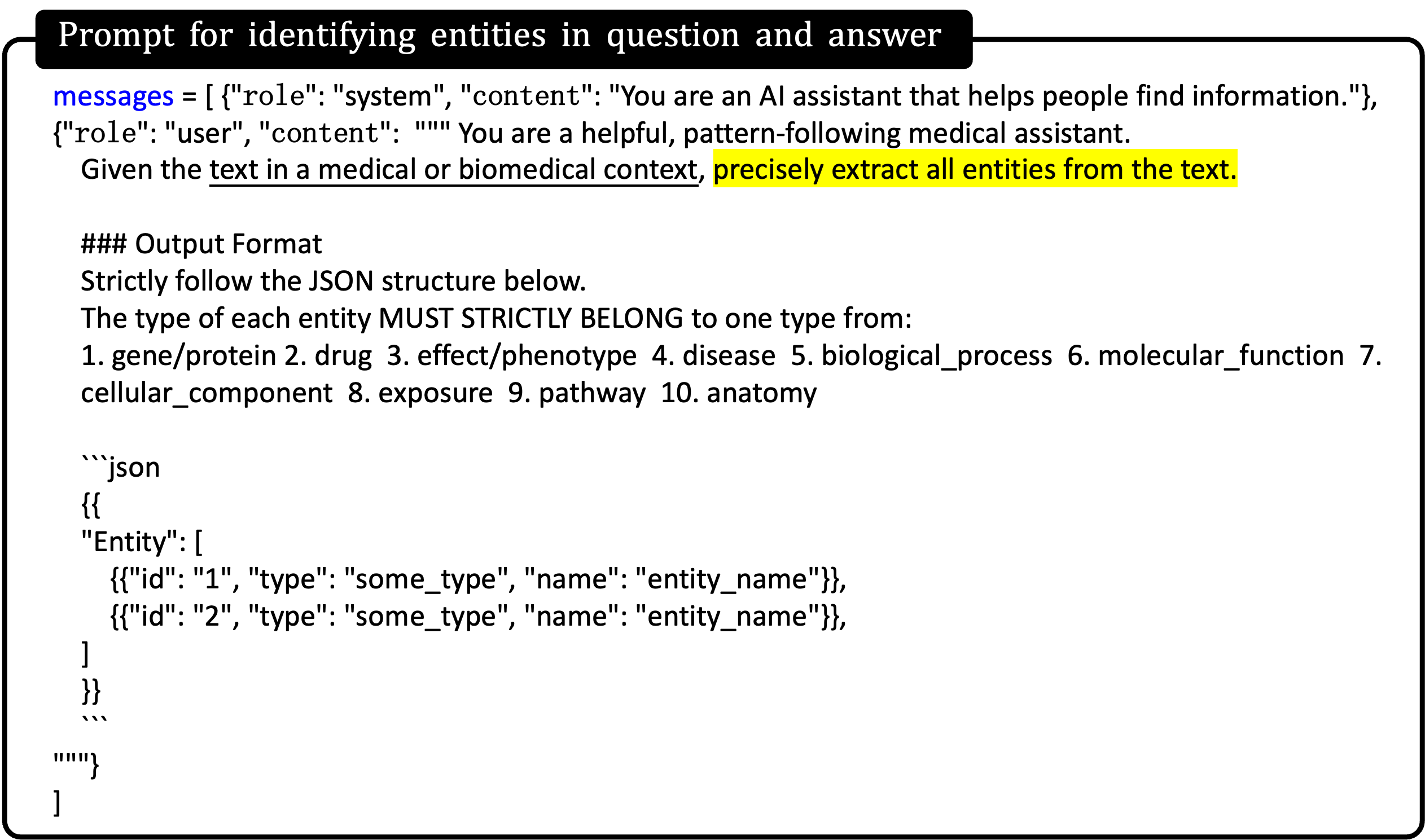} 
\caption{Prompt for identifying entities in the question and answer.}
\label{fig:ident_prompt}
\end{figure*}

\begin{figure*}[h]
\centering
\includegraphics[width=\linewidth]{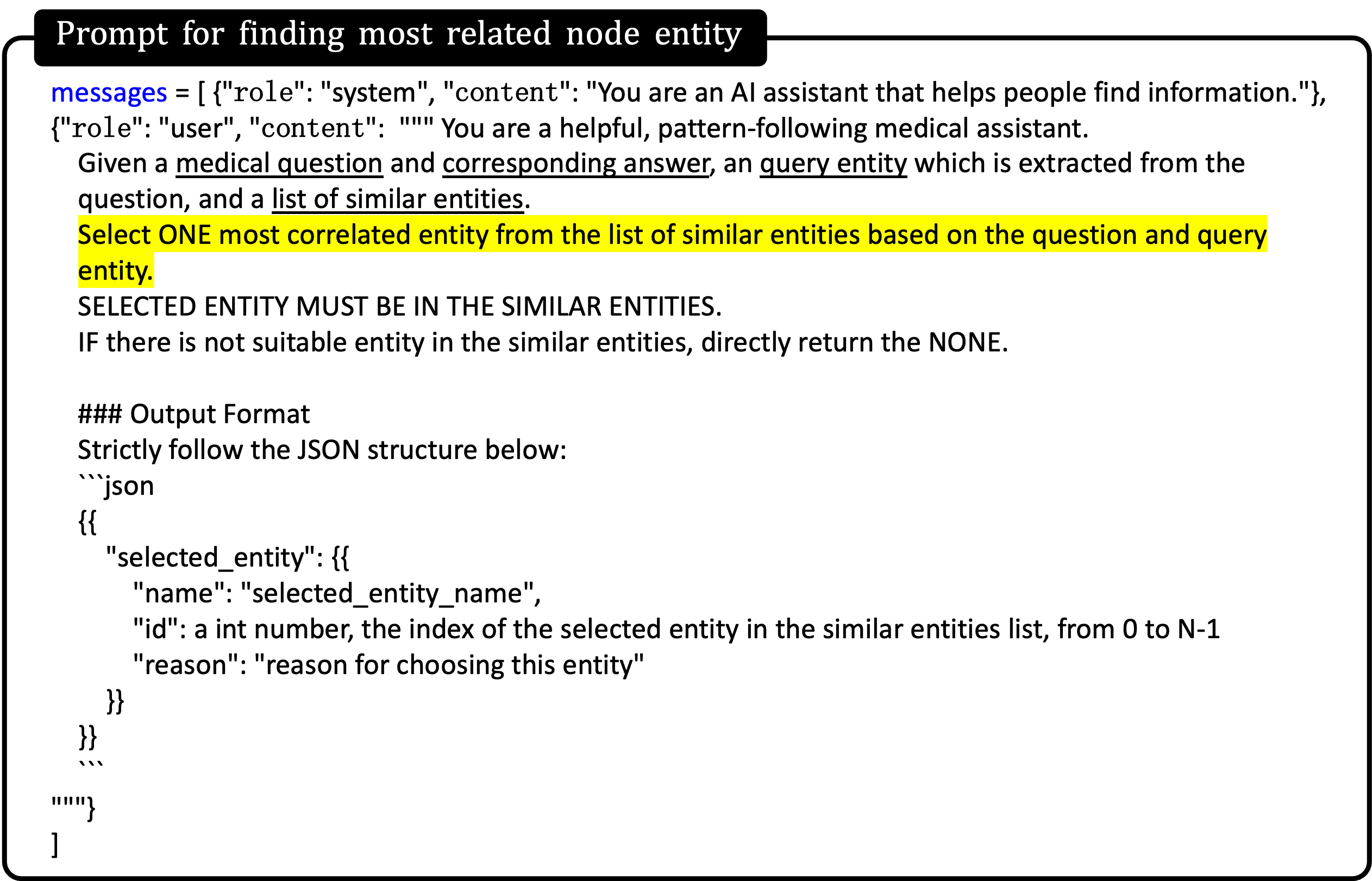} 
\caption{Prompt for selecting the most relevant nodes for extracted entities.}
\label{fig:rela_prompt}
\end{figure*}

\begin{figure*}[h]
\centering
\includegraphics[width=\linewidth]{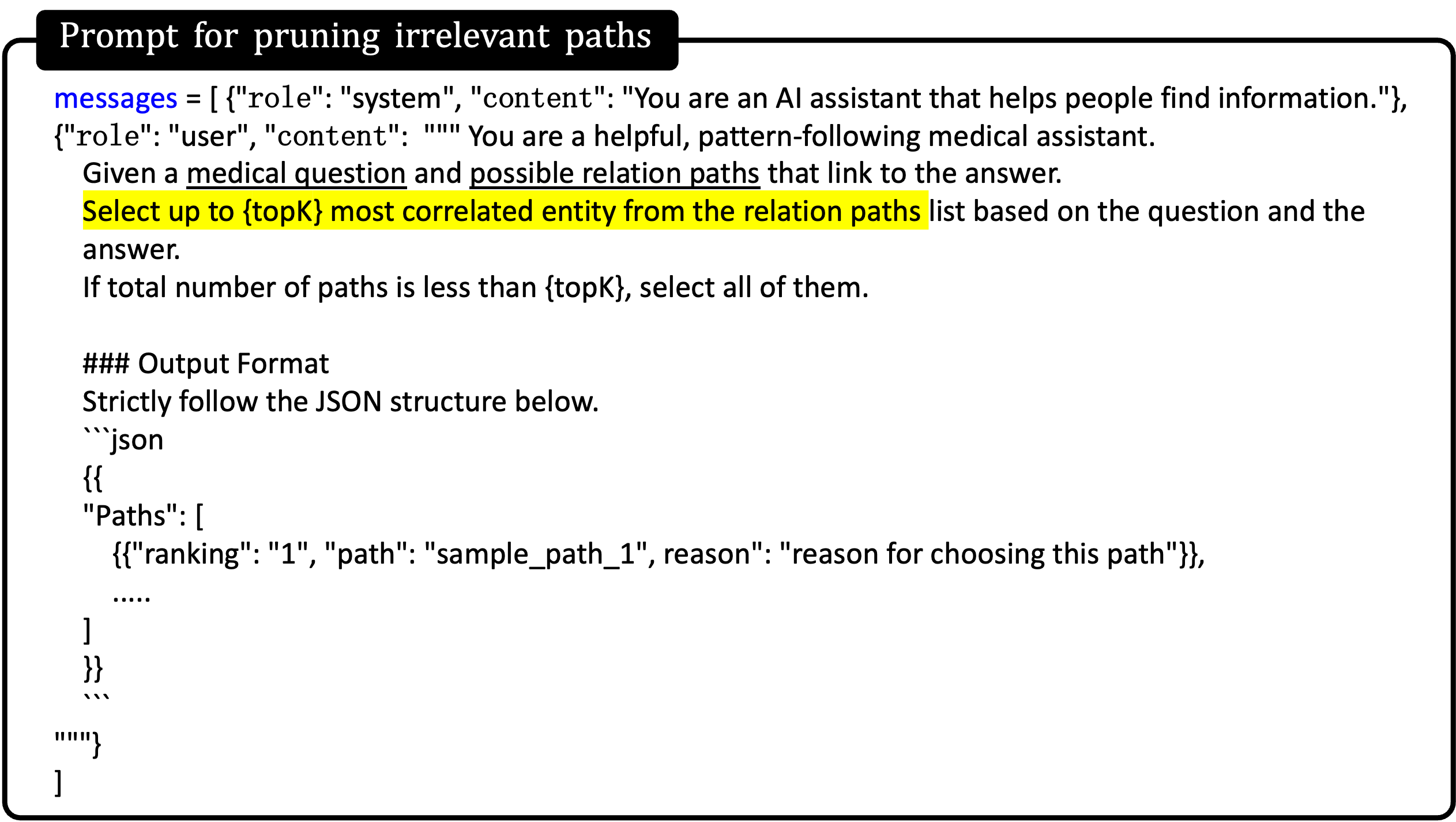} 
\caption{Prompt for pruning irrelevant paths.}
\label{fig:purne_prompt}
\end{figure*}

\begin{figure*}[h]
\centering
\includegraphics[width=\linewidth]{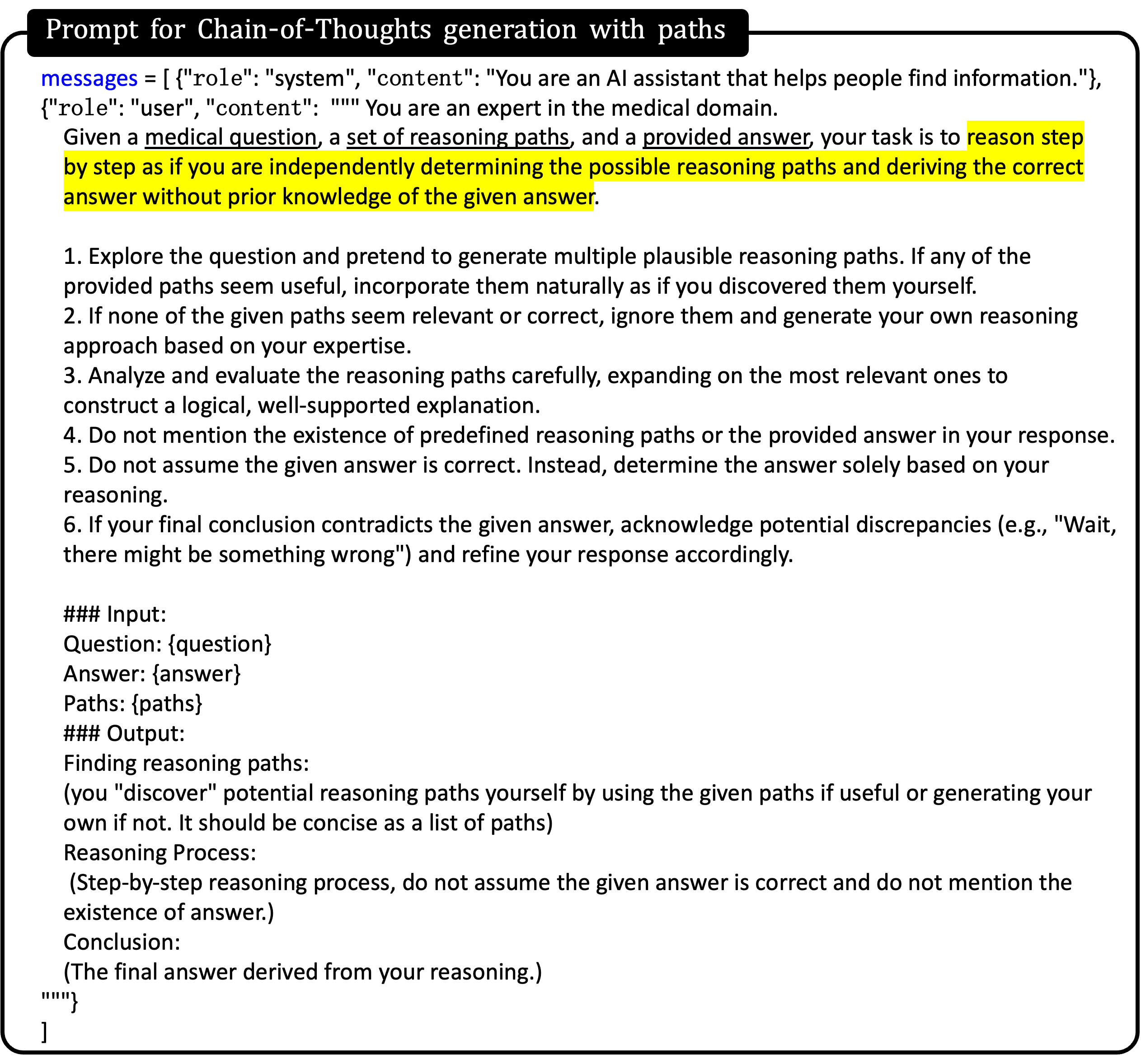} 
\caption{Prompt for Chain-of-Thoughts generation with paths.}
\label{fig:gen_prompt}
\end{figure*}

\begin{figure*}[h]
\centering
\includegraphics[width=\linewidth]{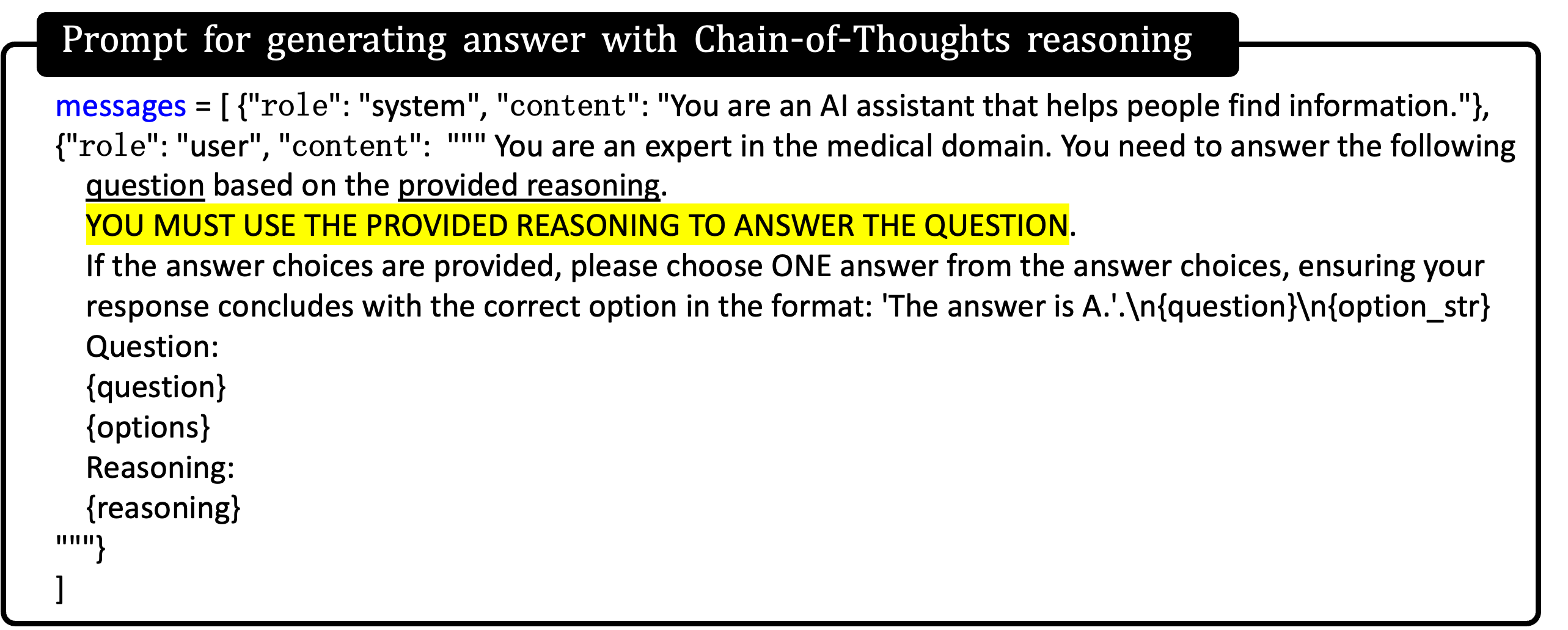} 
\caption{Prompt for generating the answer based on provided CoT data.}
\label{fig:eval_prompt}
\end{figure*}

The prompts used in our data generation pipeline are shown in the following figures. We provide details about prompts for (a) identifying entities in the question and answer (\cref{fig:ident_prompt}), (b) selecting the most relevant nodes for extracted entities (\cref{fig:rela_prompt}), (c) pruning irrelevant paths (\cref{fig:purne_prompt}), (d) CoT data generation with paths (\cref{fig:gen_prompt}), and (e) generating the answer based on provided CoT data for quality filtering (\cref{fig:eval_prompt}).

\subsection{Detailed Algorithm}
In~\cref{alg:entity_mapping}, we present our detailed algorithm for entity extraction and mapping, as outlined in~\cref{sec:entity_extraction}. First, a text embedding model is used to encode each entity \( e \in E \) and compute its similarity with the node embeddings in the \( G \). This generates a ranked list of candidate matches, from which we then extract the Top-\( K \) most similar entities to form a candidate set \( S \). Thirdly, we select the final entity from the Top-\( K \) entities, following the three matching stages: 
\\
\noindent \textbf{Stage 1 (Exact Match):} The algorithm iterates over \( S \) and checks for an exact match with \( e \). If an exact match is found $e \in S$, the corresponding node is selected.  
\\
\textbf{Stage 2 (Similarity Match):} If an exact match is not found and the top similarity score exceeds a predefined threshold \( \tau \) (set to 0.85 in our case), the most similar entity from \( S \) is selected.  
\begin{align}
    \hat{e} = \arg\max_{s_k \in S} \cos(e,s_k), ~\text{if}  ~\cos(e,s_k) > \tau
\end{align}
\\
\textbf{Stage 3 (LLM-based Selection):} If no suitable candidate is found in the above two stages, we instruct the LLM to analyze the question-answer context and the entity name to determine the most relevant node from \( S \). The selection prompt $I_{\text{select}}$ is demonstrated in \cref{fig:rela_prompt} in the Appendix.
\begin{align}
     \hat{e} &= LLM\left(S, Q, A\mid I_{\text{select}} \right),
\end{align}
Finally, we derive mapped entity sets from the graph, denoted as \( \{\hat{e}_i^Q\}_{i \in [n]} \) and \( \{\hat{e}_j^A\}_{j \in [m]} \), respectively. As illustrated in Fig.~\ref{fig:example}, besides the entities \textit{difficulty walking} and \textit{broad-based gait} which exactly match graph nodes, the entity \textit{bilateral optic disc swelling} is mapped to a similar concept, \textit{Abnormality of the optic disc}, in the knowledge graph.
\begin{algorithm}[h]
\caption{Mapping Extracted Entities to Knowledge Graph Nodes}\label{alg:entity_mapping}
\begin{algorithmic}[1]
\Require Extracted entity set \( \mathcal{E} \) from \( Q \) or \( A \), Knowledge Graph \( G \), text embedding model $f_T$, similarity threshold \( \tau \), lower case operation \texttt{lower}, prompt instruction $\mathcal{I}_s$ and LLM function \( f_{L} \).
\Ensure Mapped knowledge graph nodes for each entity in \( \mathcal{E} \)
\For{each entity \( e \) in \( E \)}
    \State \( S \gets \texttt{get\_similar\_entities}(e, G, f_T) \)
    \State \( s_{\text{top}} \gets \texttt{top\_similarity}(S) \)
    \State \( \texttt{selected} \gets \texttt{None} \)

    \For{each candidate \( c \) in \( S \)} \Comment{Stage 1: Exact Match}
        \If{\( \texttt{lower}(e) = \texttt{lower}(c) \)}
            \State \( \texttt{selected} \gets c \)
            \State \textbf{break}
        \EndIf
    \EndFor

    \If{\( \texttt{selected} = \texttt{None} \) and \( s_{\text{top}} > \tau \)} \Comment{Stage 2: Similarity Match}
        \State \( \texttt{selected} \gets S[0] \)
    \EndIf

    \If{\( \texttt{selected} = \texttt{None} \)} \Comment{Stage 3: LLM-based Selection}
        \State \( \texttt{selected} \gets f_{L}(Q,A, e, S | \mathcal{I}_s) \)
    \EndIf
    \State \texttt{Map } \( e \) \texttt{ to node } \(\texttt{selected}  \) $\hat{e}$
\EndFor
\end{algorithmic}
\end{algorithm}

\subsection{Detailed comparison between \texttt{MedReason}-8B and Huatuo-o1-RL-8B.}
\begin{figure*}[h]
\centering
\includegraphics[width=\linewidth]{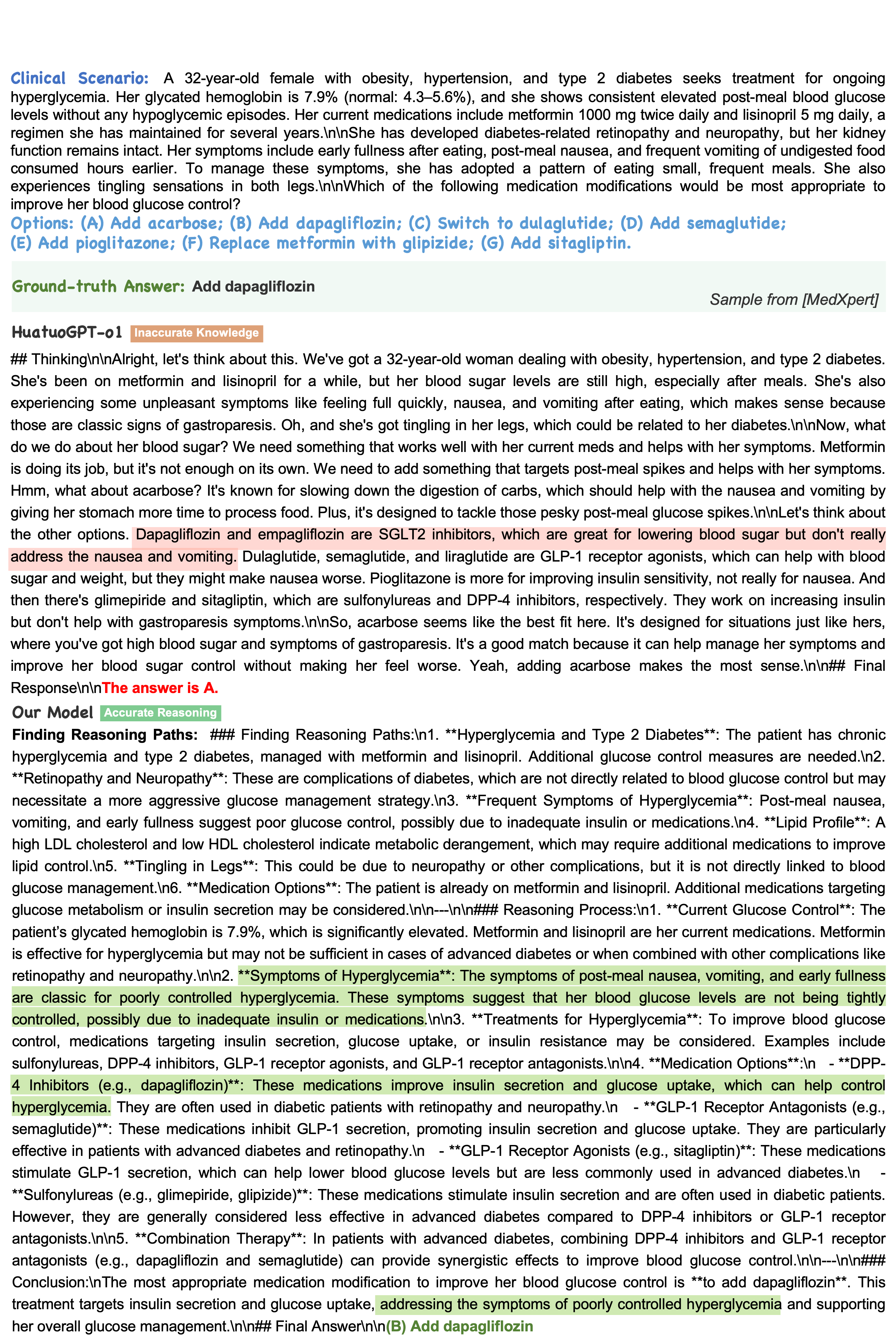} 
\caption{Full comparison between \texttt{MedReason}-8B and Huatuo-o1-RL-8B.}
\label{fig:case_appendix}
\end{figure*}

In~\cref{fig:case_appendix}, we present a comprehensive comparison of \texttt{MedReason}-8B and Huatuo-o1-RL-8B in tackling complex clinical issues. Huatuo-o1-RL-8B delivered an incorrect answer due to overlooking the link between vomiting and hyperglycemia, whereas our \texttt{MedReason}-8B accurately identified the correct answer, grounded in factual reasoning.

\end{document}